\documentclass[10pt,twocolumn,letterpaper]{article}

\usepackage[pagenumbers]{cvpr} 

\usepackage{graphicx}
\usepackage{amsmath}
\usepackage{amssymb}
\usepackage{booktabs}

\usepackage[pagebackref,breaklinks,colorlinks]{hyperref}

\usepackage[capitalize]{cleveref}
\crefname{section}{Sec.}{Secs.}
\Crefname{section}{Section}{Sections}
\Crefname{table}{Table}{Tables}
\crefname{table}{Tab.}{Tabs.}

\def\cvprPaperID{*****} 
\def\confName{CVPR}
\def\confYear{2022}

\begin{document}

\title{VROOM - Visual Reconstruction over Onboard Multiview}

\author{Yajat Yadav\\
{\tt\small yajatyadav@berkeley.edu}
\and
Varun Bharadwaj\\
{\tt\small varunbharadwaj@berkeley.edu}
\and
Jathin Korrapati\\
{\tt\small jkorr@berkeley.edu}
\and
Tanish Baranwal\\
{\tt\small tekotan@berkeley.edu}
}

\maketitle

\begin{abstract}
   We introduce VROOM, a system for reconstructing 3D models of Formula 1 circuits using only onboard camera footage from racecars. Leveraging video data from the 2023 Monaco Grand Prix, we address video challenges such as high-speed motion and sharp cuts in camera frames. Our pipeline analyzes different methods such as DROID-SLAM, AnyCam, and Monst3r and combines preprocessing techniques such as different methods of masking, temporal chunking, and resolution scaling to account for dynamic motion and computational constraints. We show that Vroom is able to partially recover track and vehicle trajectories in complex environments. These findings indicate the feasibility of using onboard video for scalable 4D reconstruction in real-world settings. The project page can be found at https://varun-bharadwaj.github.io/vroom, and our code is available at https://github.com/yajatyadav/vroom.
\end{abstract}

\section{Introduction}
\label{sec:intro}
\subsection{F1}
In this project, our aim is to design an algorithm to reconstruct a Formula 1 circuit from the onboard cameras of racecars. The dataset we use is the onboard of a race with 20 cars racing around a 2.074-mile-long circuit in the principality of Monaco known as the Monaco Grand Prix.
\begin{figure}[h]
    \centering
    \includegraphics[width=1\linewidth]{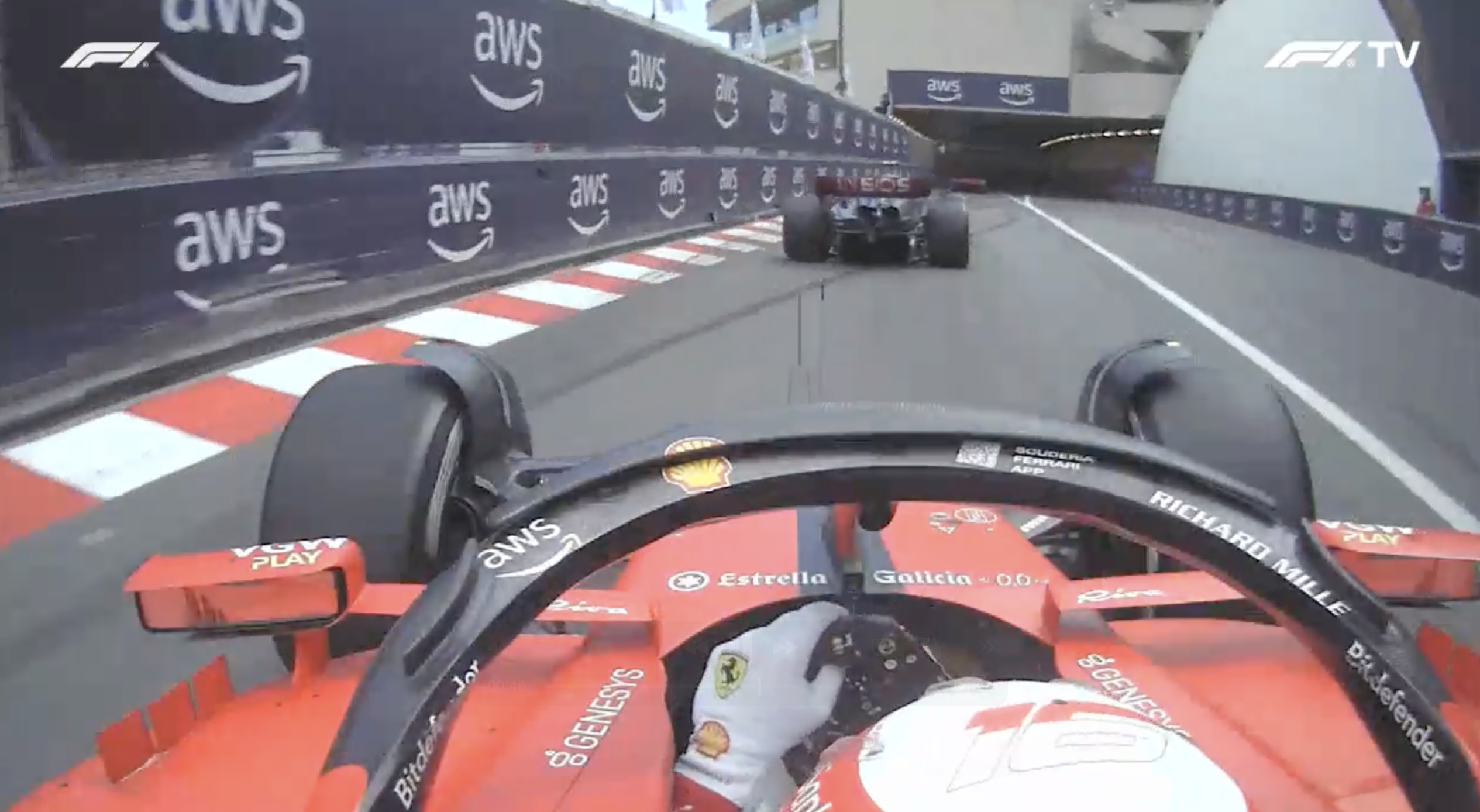}
    \caption{Example onboard video}
    \label{fig:enter-label}
\end{figure} \\
The main challenges we encounter when using Formula 1 video data are high-speed 3-D reconstruction and 3-D reconstruction from multiple views. Formula 1 cars are known for their high cornering speed, which makes the problem of 3-D reconstruction difficult because the cars move quickly through high-detail twists in the circuit. This is challenging because most 3-D reconstruction methods depend in some way on optical flow estimation, which is very difficult in scenes with dynamic motion.
\begin{figure}[h]
    \centering
    \includegraphics[width=1\linewidth]{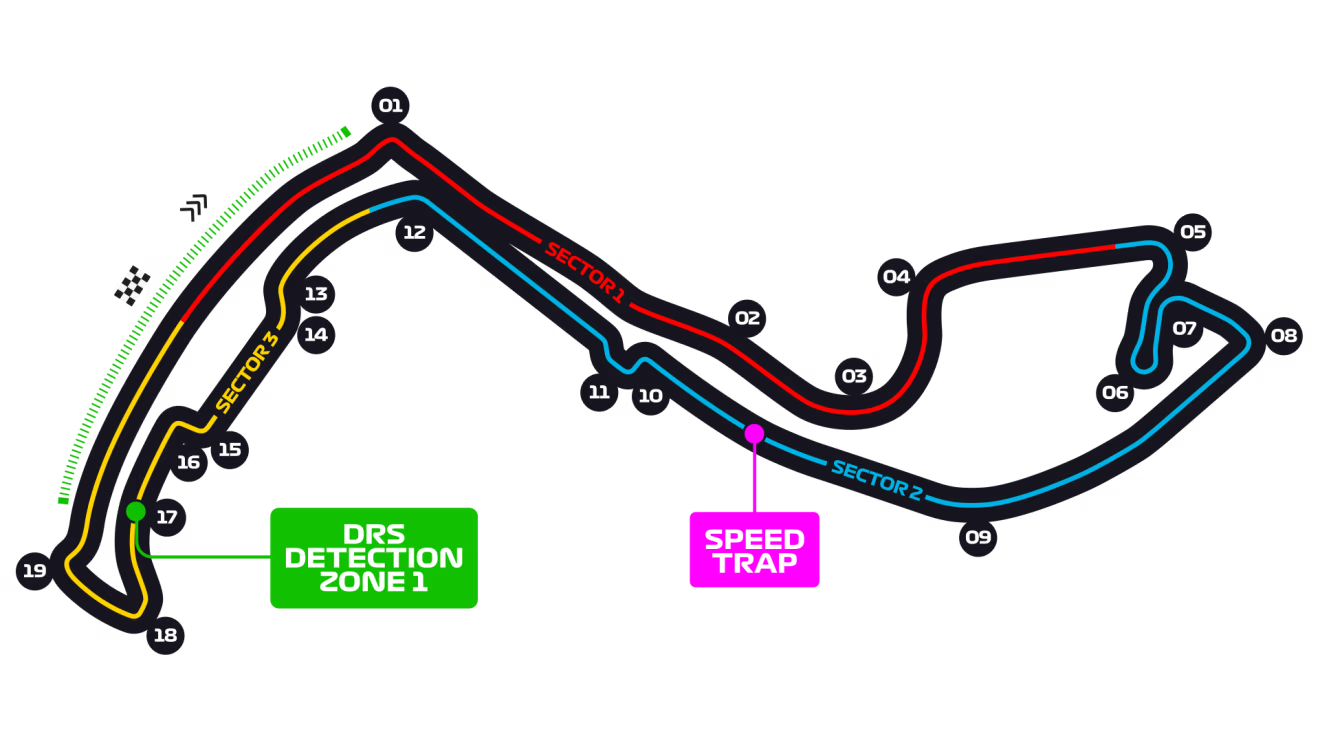}
    \caption{Monaco Track Layout}
    \label{fig:enter-label}
\end{figure} \\
The specific reason we chose to do the Monaco Formula 1 track is because of the small distance between the cars and the walls. This yields a higher quality optical flow because of the closer distance between the moving objects and the camera. Furthermore, the Monaco track has a lot of tight twists and turns, especially in the second sector s (turns 5-12), really testing how well the model learns the 3-D reconstruction from the images.
\subsection{SLAM}
Simultaneous Localization and Mapping (SLAM) is a key part in many robotic and autonomous systems, which requires the system to predict/learn both its surroundings and where it is within its environment. In our project, we try to calculate the track layout as well as the location of the car on the track. We are trying to calculate the racing line, the path that the car takes as it makes it way through the track. Figure 3 shows a sample racing line that F1 cars usually try to follow. As we learn the camera extrinsics, we learn the racing line that the cars must take in order to minimize distance traveled while maximizing speed.
\begin{figure}[h]
    \centering
    \includegraphics[width=1\linewidth]{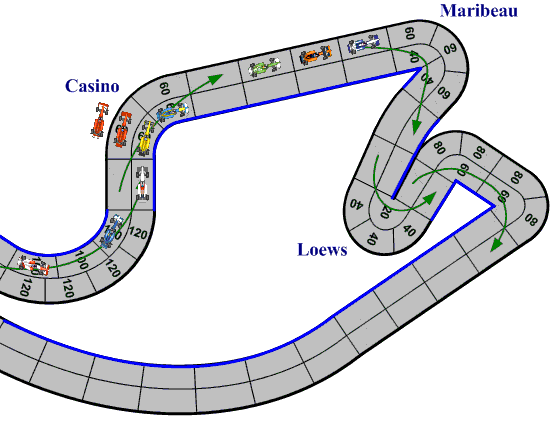}
    \caption{Racing Line}
    \label{fig:enter-label}
\end{figure}
\subsection{3D Reconstruction}
Along with SLAM, we perform a 3-D reconstruction to create a dense 3-D point cloud of the entire track. To do this we combine the depth maps + estimated position-wise point cloud from the SLAM outputs along with the estimated camera extrinsics. This dense 3-D reconstruction can be combined with the estimated point clouds from all other onboard cameras to develop a 3-D reconstruction of all 20 cars and the track through time.
\section{Related Work}

Traditional SLAM and Structure-from-Motion (SfM) pipelines have provided reliable camera tracking and 3D scene reconstruction for decades, leveraging geometric priors and hand-engineered features. However, the reliance on well-textured scenes, static environments, and calibrated cameras limits their performance in unconstrained scenarios. Recent years have witnessed a transition toward learning-based SLAM systems that offer improved robustness, generalization, and scalability. We survey this evolution, with a focus on monocular and multi-camera SLAM pipelines, culminating in the latest dynamic-scene and camera-agnostic models.

\subsection{Classical and Early Learning-Based SLAM.} Foundational systems like ORB-SLAM~\cite{mur2015orb, campos2021orb} and LSD-SLAM~\cite{engel2014lsd} popularized keyframe-based mapping, bundle adjustment, and direct methods. ORB-SLAM3 extended this pipeline to support multi-camera and visual-inertial setups with remarkable accuracy. In parallel, learning-based methods such as PoseNet~\cite{kendall2015posenet} and SfM-Learner~\cite{zhou2017unsupervised} introduced direct pose and depth regression from images using deep networks, enabling unsupervised learning from video via view synthesis. However, early learned approaches struggled with scale ambiguity, drift, and generalization.

\subsection{Differentiable SLAM.} A breakthrough came with DROID-SLAM~\cite{teed2021droid}, which demonstrated that classical SLAM principles could be embedded in an end-to-end differentiable framework. DROID-SLAM introduces a dense, differentiable bundle adjustment layer that iteratively refines pose and depth via learned updates. Its recurrent optimization backbone, inspired by optical flow methods like RAFT~\cite{teed2020raft}, enabled robust SLAM performance across monocular, stereo, and RGB-D modalities without retraining. DROID-SLAM outperformed traditional pipelines on static benchmarks such as TUM RGB-D and TartanAir, establishing a new standard for learned SLAM.
\subsection{Dynamic Scene SLAM.} While DROID-SLAM is robust in static environments, its assumptions break down in the presence of dynamic objects or low-parallax motion. MegaSaM~\cite{li2024megasam} addresses these limitations by introducing dynamic-scene modeling within the SLAM loop. It employs monocular depth priors, per-pixel motion probability masks, and a learned curriculum that transitions from static to dynamic training. Furthermore, it jointly optimizes intrinsic camera parameters at inference time, allowing for uncalibrated, casually captured video. MegaSaM demonstrates state-of-the-art pose and depth estimation on challenging dynamic datasets, outperforming optimization-heavy pipelines like Casual-SfM.

\subsection{Feed-Forward 4D Reconstruction.} Complementary to optimization-based methods, MonST3R~\cite{wang2024monst3r} proposes a feed-forward pipeline that directly estimates 3D pointmaps per video frame, adapting the static-scene architecture DUSt3R~\cite{chen2024dust3r} to dynamic settings. Rather than explicitly modeling object motion, MonST3R predicts per-frame 3D structure aligned in a common coordinate system, with temporal consistency enforced via a lightweight global alignment step. Trained on a modest corpus of dynamic scenes, MonST3R offers a compelling trade-off between simplicity, speed, and robustness, producing temporally coherent 3D reconstructions even under object motion.

\subsection{Generalizing Across Camera Models.} Most SLAM pipelines assume calibrated pinhole cameras, limiting deployment in real-world applications with wide-FOV or unknown intrinsics. UniK3D~\cite{piccinelli2025unik3d} tackles this by learning a spherical 3D representation coupled with a camera-agnostic ray encoding via spherical harmonics. This disentangles geometry prediction from camera intrinsics, enabling monocular 3D reconstruction across fisheye, panoramic, and perspective lenses without calibration. While not a full SLAM system, UniK3D's universal camera model offers an important building block for future multi-camera SLAM and mapping frameworks.

\section{Methods}
\label{sec:formatting}

\subsection{Data}
Using F1 TV, we first obtained 20 onboard videos for each driver from the 
2023 Monaco Grand Prix race. We decided to focus on the Monaco track for it's high fidelity features, like the changing background buildings, as well as varied turns and elevations in the track. This track would serve as a robust benchmark ensuring our method can succesfully reconstruct F1 tracks as well as the cars' motion in general.
\subsection{Preprocessing}
\subsubsection{Downsampling}
The original video data consisted of 1280x720, 50 FPS video, with each about 80 seconds long. As we were limited by compute and hoped to iterate on our method faster, we downsampled the final resolution to 512x144, as well as lowered the FPS to 24 FPS. By testing against several, short 5 second chunks of the original video, we found this downsampling of resolution and FPS had marginal imapct on the final reconstruction quality.
    \subsubsection{Masking}
    The onboarding camera videos, as seen in Figure 1 were what we initially used as data input with no changes. However, the car remains stationary throughout the entire race with respect to the camera, and this interfered with all the data-based SLAM methods that we tried. A common issue was that the learned 3D reconstruction would end up having a red "cylinder" throughout the track looking like the car.
    
    To counteract this, we implemented a masking technique that would block a portion of the frame (i.e. the top $30\%$ or top $50\%$). The intuition was removing the stationary car from the original input data sequence might help reconstruct the motion better relative to the F1 track which would improve results.

    We also experimented with other masking methods, such as masking just the car (Figure 4), carving out specific masks, or masking higher/lower fraction of the frame, and found that masking out the bottom half of the frame (Figure 5) performed the best. Once we discovered this, we added this masking to our list of preprocessing methods.   
    \begin{figure}[h]
    \centering
    \begin{minipage}[b]{0.45\linewidth}
        \centering
        \includegraphics[width=\linewidth]{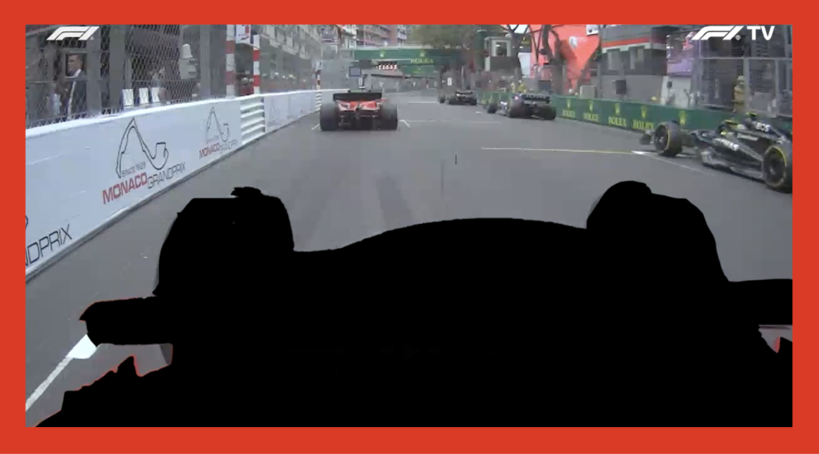}
        \caption{Masking Car}
    \end{minipage}
    \hfill
    \begin{minipage}[b]{0.45\linewidth}
        \centering
        \includegraphics[width=\linewidth]{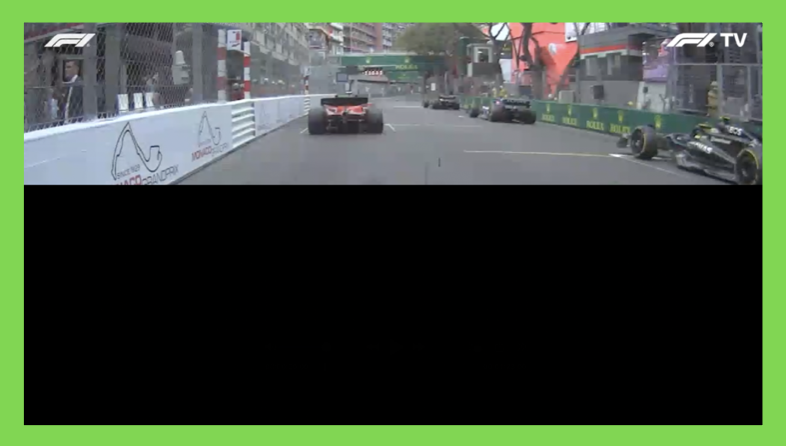}
        \caption{Masking Bottom Half}
    \end{minipage}
\end{figure}
    \subsubsection{Video Chunking}
    Finally, because we were limited by our compute, we had a tradeoff between feeding longer videos at the expense of lower FPS in the various methods we tried. With experimentation, we found lowering the FPS too much (lower than 12) would lead to terrible reconstruction, likely due to the cars' extremely fast motion in the world. Thus, we kept our 24 FPS video but chunked it into overlapping segments. After some experimentation, we settled on about 5 second chunks, with 1 frame of overlap.
    So, our final pipeline consisted of processing these video chunks individually and then combining each of the learned camera motions and point clouds into one (described in section 3.5).
    \subsubsection {Smarter Chunking}
    While the naive chunking approach above (just taking 5 seconds at a time) did improve the quality of our reconstruction, the point cloud would often deteriorate near the turns. Upon analyzing the outputs, we found this to be the issue when a turn would get cut off and be present in two contiguous chunks. Thus, informed with this, we revised our chunking algorithm to only split the video on straight segments. This way, we ensured that each turn was fed into our method with enough context so the 3D reconstruction could accurately capture the curvature of the turn. 

\subsection{Race Reconstruction}
With our processed data, we tried a few different methods from our literature review to obtain both 1) 3D reconstructions of the F1 track, and 2) the camera motion in the track. In this project, we mainly focused on getting these methods to work using one car's onboard video; combining multiple cars' viewpoints to further refine the 3D reconstruction and car movements is a future direction that we hope to explore.

\subsubsection{Droid-SLAM}
    The first method we tried to adapt to was Droid-SLAM. Droid-SLAM estimates camera poses and scene geometry from videos and utilizes the principles of feature extraction with a CNN and uses recurrent bundle adjustment to optimize on the data. We thought using Droid-SLAM would help us in estimating the relative motion of the frame to generate the trajectory of the car to rebuild the track. Our intention was to first use Droid-SLAM as a baseline method to see how well we could generate the reconstruction based on our data. However, we found that due to our compute not having GUI support it was difficult to visualize reconstruction, even for the examples given in the official repository. Droid-SLAM from the limited examples we have also does not reproduce motion, but just the static elements of the track. Before figuring out the visualization problem, we got AnyCam and Monst3r working, so we ditched our efforts with Droid-SLAM.

\subsubsection{AnyCAM}
    The next method we tried adapting to our problem was AnyCam. AnyCam employs a transformer-based network to output per-frame camera poses and intrinsics. Being trained on large-scale real-world camera movements and being much faster at test-time than iterative SLAM methods, AnyCam seemed to be a promising candidate. 
\subsubsection{Monst3r}
    The best results we got were using Monst3r, a library developed to estimate the geometry of scenes with motion. Monst3r is an addition to Dust3r that computes time-indexed point cloud and camera motion given a video as an input. The main issues we faced while using Monst3r was its high memory usage. We were unable to load the entire video (even at 1fps) using an NVIDIA H200 132GB VRAM GPU, and saw subpar results with a longer video but too low of a FPS (Figure 6). In order to fix this, we had to apply the masking, downsampling, and chunking described in the preprocessing section. 

    After using Monst3r to process each chunk of video, we then utilized the overlap between chunks in order to transform one chunk's camera extrinsics into another camera's reference frame. By repeating this process repeatedly with all chunks, we were able to "stitch" together the trajectories and obtain a single camera trajectory. Furthermore, we adjusted the point cloud rendering to use these transformed extrinsics to reconstruct the entire 3D scene for the F1 track. As desired, the end result was a 3-D model for the entire track, as well as the car's trajectory.

    We next aimed to address the problem of "closing the loop", i.e. ensuring that combining together each chunk's reconstruction still leads to a global reconstruction consistent with the real world. To accomplish this, we began extending Monst3r to instead process the video in chunks, and then perform bundle adjustment across chunks in order to prevent each reconstruction from slightly drifting off. Since Monst3r builds a graph of frames and then performs bundle adjustment with pairs of frames with an edge between them, we modified Monst3r to instead do the following: for each chunk, we add edges between frames in the chunk and perform bundle adjustment to refine the chunk's reconstruction. Then, we remove these edges and randomly add edges between frames in this chunk and frames in other chunks, and then perform bundle adjustment with these pairs of frames. The idea with this approach is to alternate between refining the local, chunk-level reconstruction and refining the reconstruction's alignment with the global-level reconstruction. By repeating this alternating process multiple times for each chunk, we hoped that the "drift-off error" we were noticing happening between distant chunks' reconstructions would be mitigated. However, we were unable to finish up this modification and thoroughly test this in time, and this would be an interesting future direction to ensure that our chunk-wise 3D reconstruction approach produces a globally coherent reconstruction.
    
    \begin{figure}[h]
    \centering
    \includegraphics[width=0.5\linewidth]{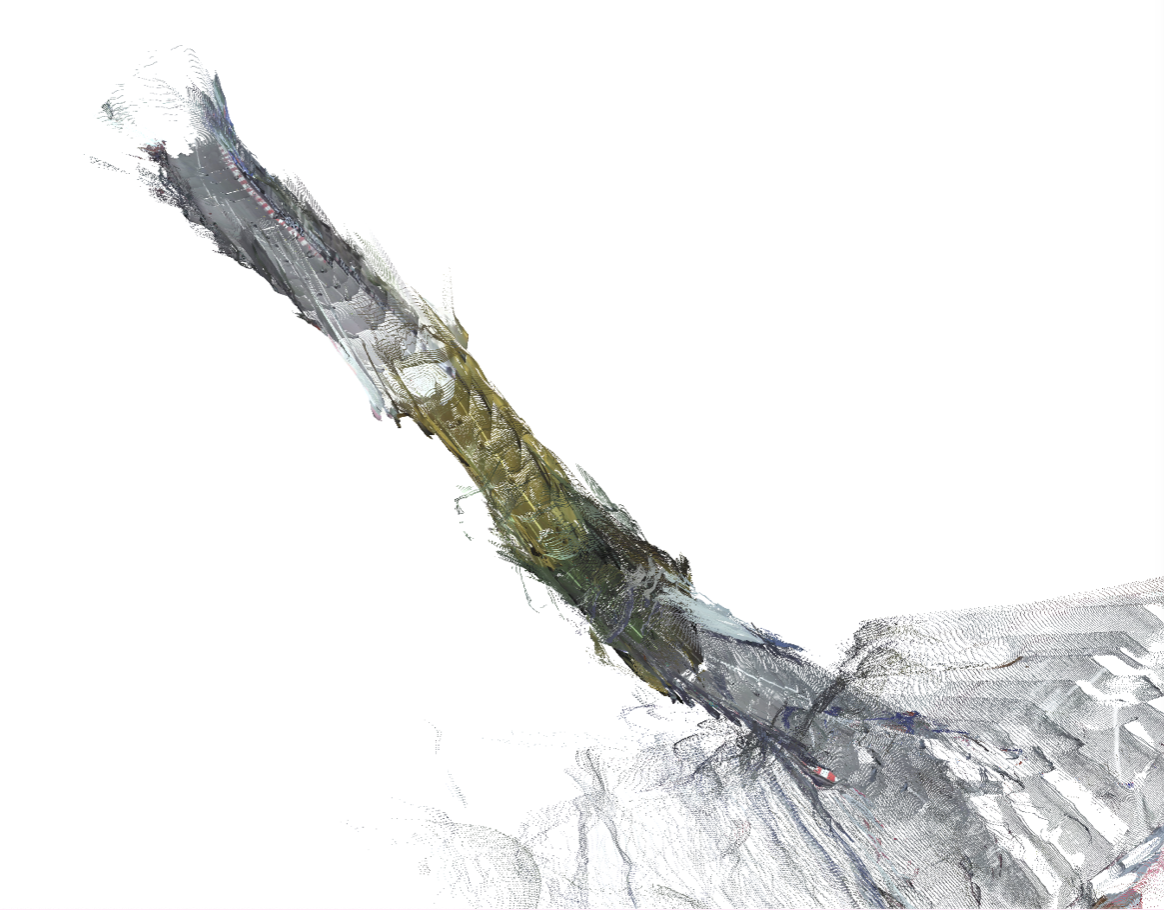}
    \caption{Poor Track reconstruction with 3 FPS}
\end{figure}
\section{Results}
\label{sec:formatting}

\subsection{AnyCAM}
We found that AnyCAM was not a robust SLAM solution in practice. It performed poorly not only on our F1 dataset, but also on dashcam-style footage resembling the examples presented in the original paper. Despite extensive sanity checks, including testing on clean, front-facing driving sequences, AnyCAM consistently failed to recover plausible camera trajectories. These failures suggest that the method may be sensitive to scene content or initialization, and raise concerns about the reproducibility of its reported performance. Figures 7 and 8 illustrate a significant mismatch between AnyCAM’s predicted camera motion and the ground truth trajectory, particularly around turns.\begin{figure}[h]
    \centering
    \begin{minipage}[b]{0.45\linewidth}
        \centering
        \includegraphics[width=\linewidth]{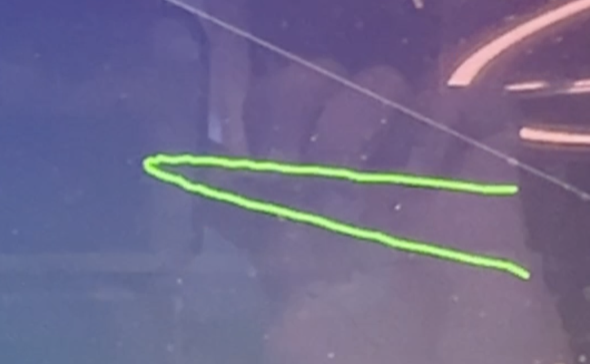}
        \caption{Predicted Camera Motion by AnyCam}
        \label{fig:anycam-pred}
    \end{minipage}
    \hfill
    \begin{minipage}[b]{0.45\linewidth}
        \centering
        \includegraphics[width=0.7 \linewidth]{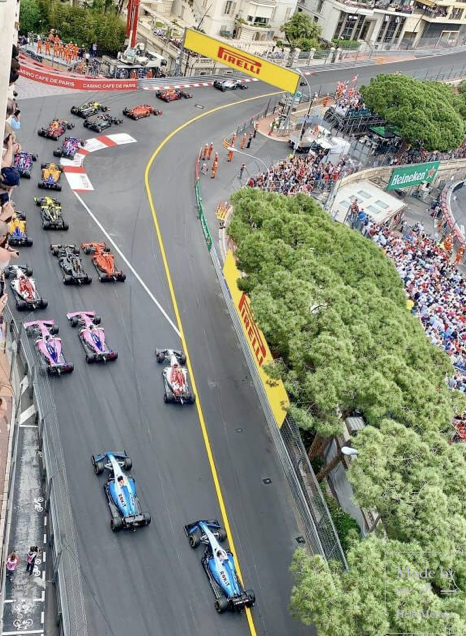}
        \caption{Ground Truth Turn}
        \label{fig:anycam-gt}
    \end{minipage}
\end{figure}

\subsection{Monst3r}
    Using the preprocessing and the chunking that we described, we were able to get the final circuit map that we have shown in figure 9. As is clear, the track does not loop back around, and there is still significant work to be done in order to  "close the loop".
    However, each individual segment's reconstruction was very close to the ground truth map, as seen in Figure 10 for turns 3, 4, and 5.

    \begin{figure}[h]
    \centering
    \includegraphics[width=0.5\linewidth]{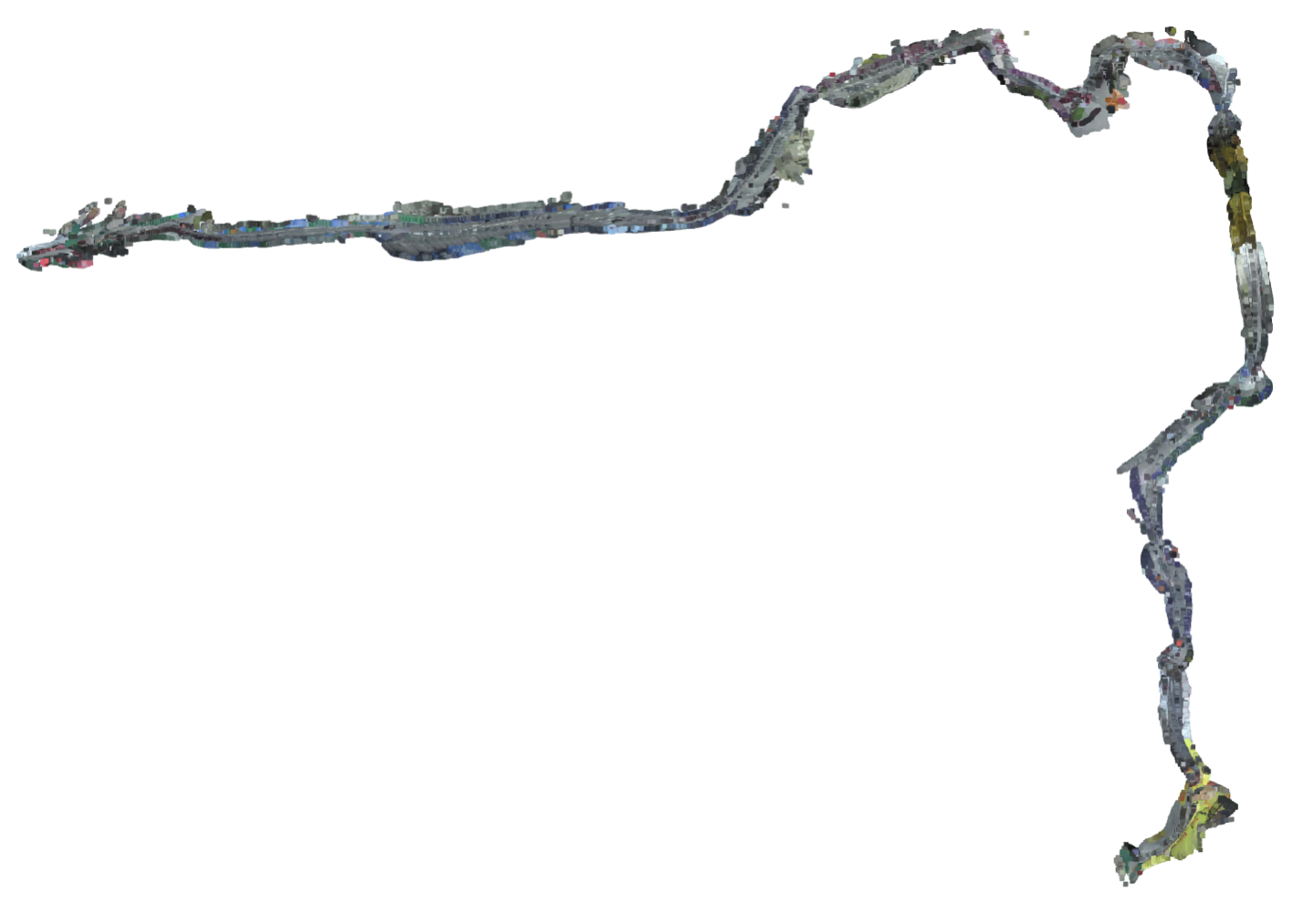}
    \caption{Full Track Reconstruction using Monst3r}
\end{figure}

    \begin{figure}[h]
    \centering
    \includegraphics[width=0.5\linewidth]{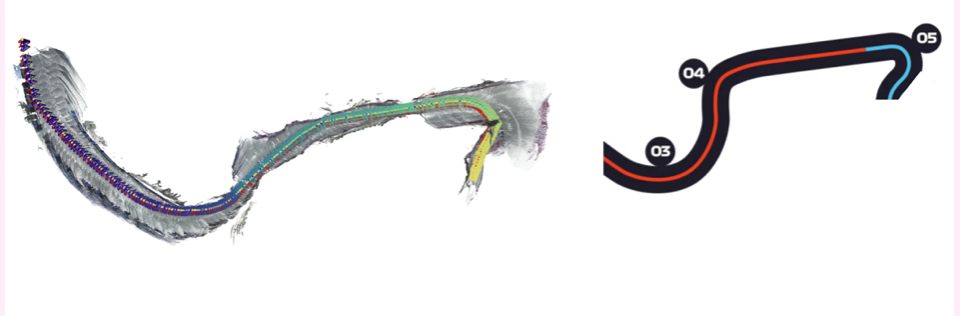}
    \caption{Turns 3/4/5 Prediction and Ground Truth}
\end{figure}

\section{Conclusion}

Our outputs from Monst3r show significant progress towards creating a 3-D reconstruction of the first lap of the 2023 Formula 1 Monaca Grand Prix. However, there is still significant progress to be made to fully fix the outputs.

\subsection{Future Improvements}
The main improvements that we were working on was to make the global reconstruction better. The main thing we were trying to do was figure out an efficient keyframe sampling method that we could then use to do global bundle adjustment. We began implementing a 2 stage process where we loop between intrachunk alignment, global bundle adjustment with keyframes only (due to Monst3r's high memory usage), followed by more iterations of intrachunk alignment and global adjustment. 
\\ \\
Furthermore, we have onboard cameras from more than just 1 camera. If we can extend this method to track the poses and learn the point clouds using multiple views. Doing this would allow us to generate a reconstruction of not just the track but also the 20 cars. Furthermore, if we had multiple views we could get rid of the masking as other views would be able to tell that the car was not a static object on the track. This would also be valuable as no masking would mean we get to use the high-quality optical flow information of the ground right next to the car to get stronger estimates of the cameras' trajectories.

{\small
\bibliographystyle{ieee_fullname}
\bibliography{egbib}
}

\newpage
\appendix
\section{Additional Figures}

\begin{table}[h]
    \centering
    \begin{tabular}{cc}
        \includegraphics[width=0.45\linewidth]{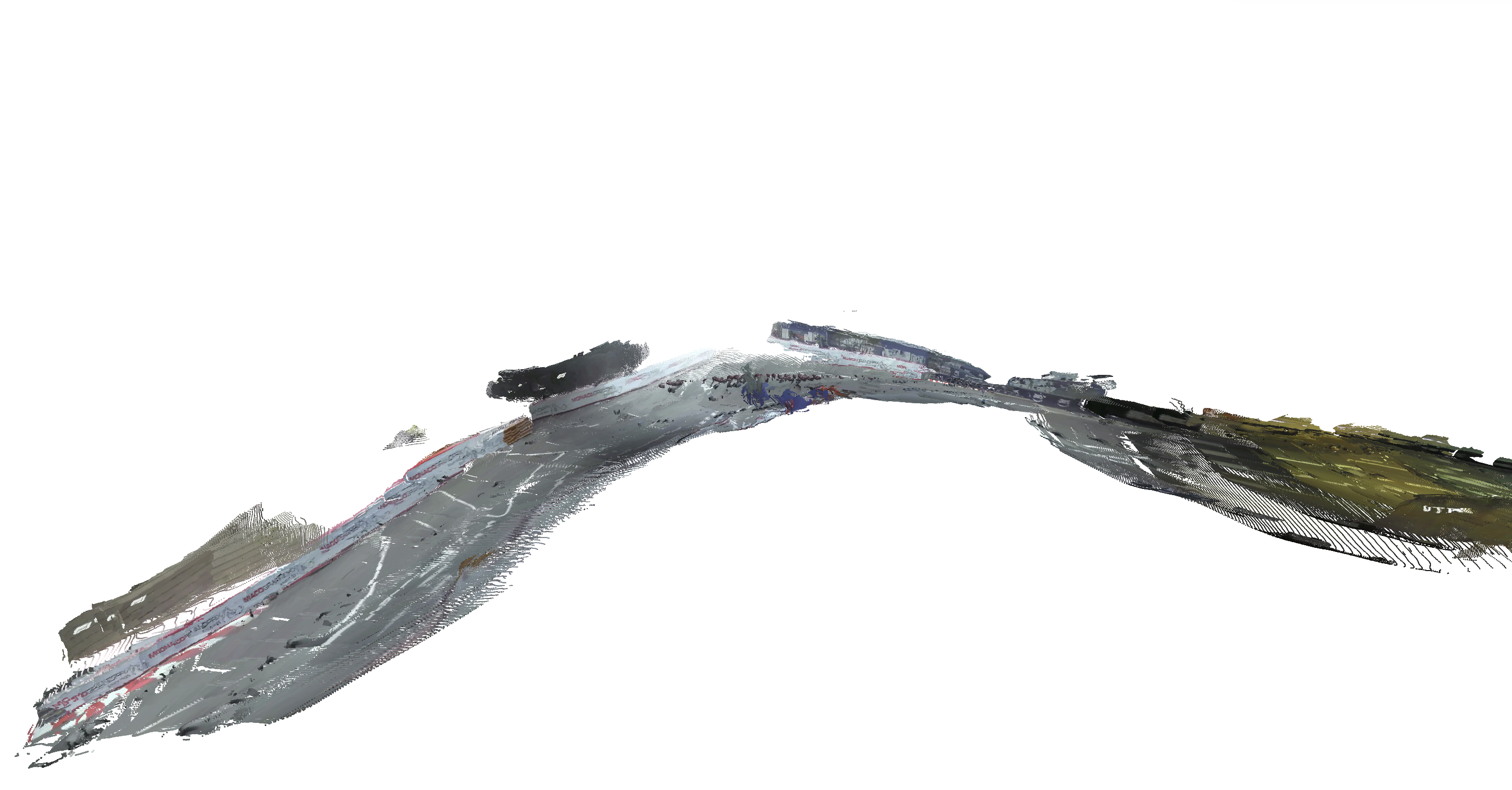} & 
        \includegraphics[width=0.45\linewidth]{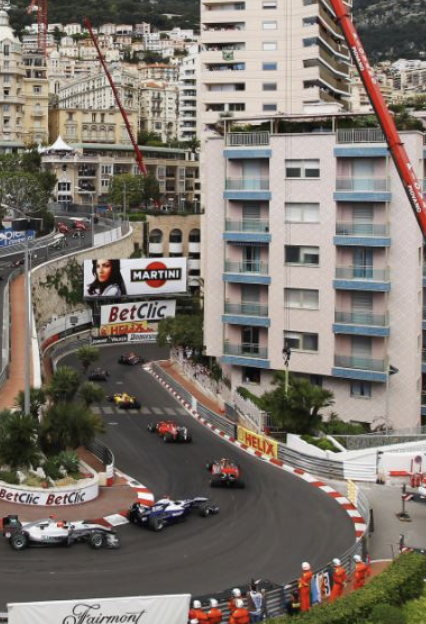} \\
        Turn 8 Reconstruction & Turn 8 Ground Truth \\
        \includegraphics[width=0.45\linewidth]{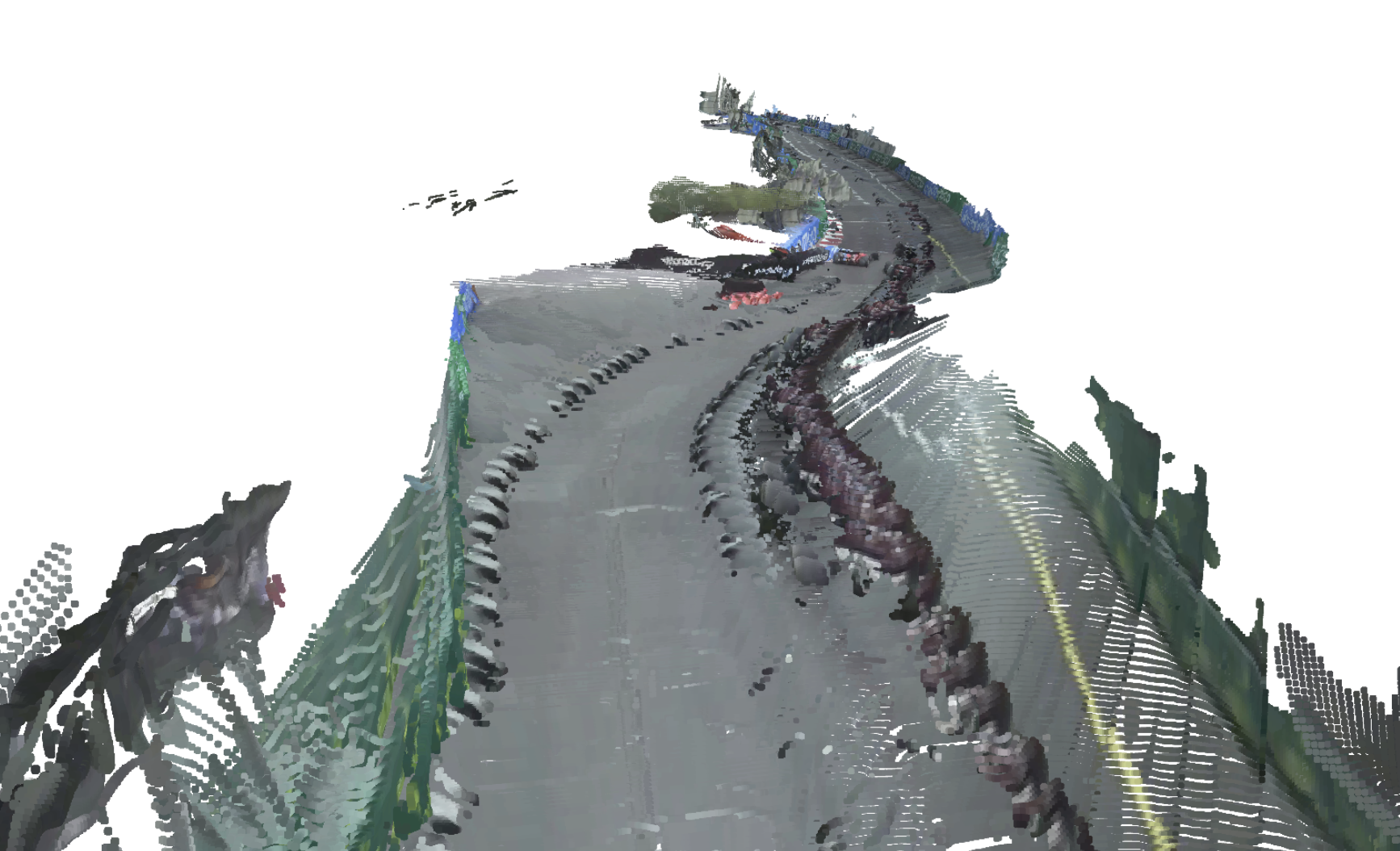} & 
        \includegraphics[width=0.45\linewidth]{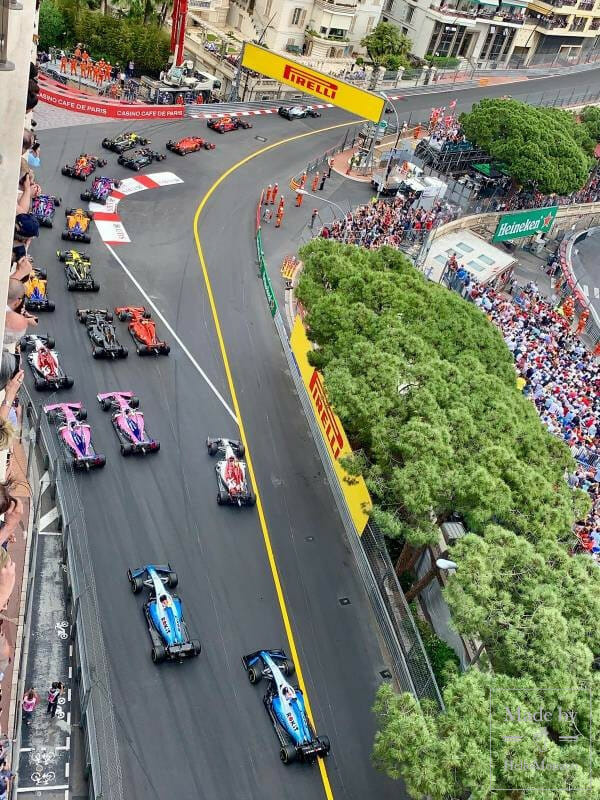} \\
        Turn 1 Reconstruction & Turn 1 Ground Truth \\
        \includegraphics[width=0.45\linewidth]{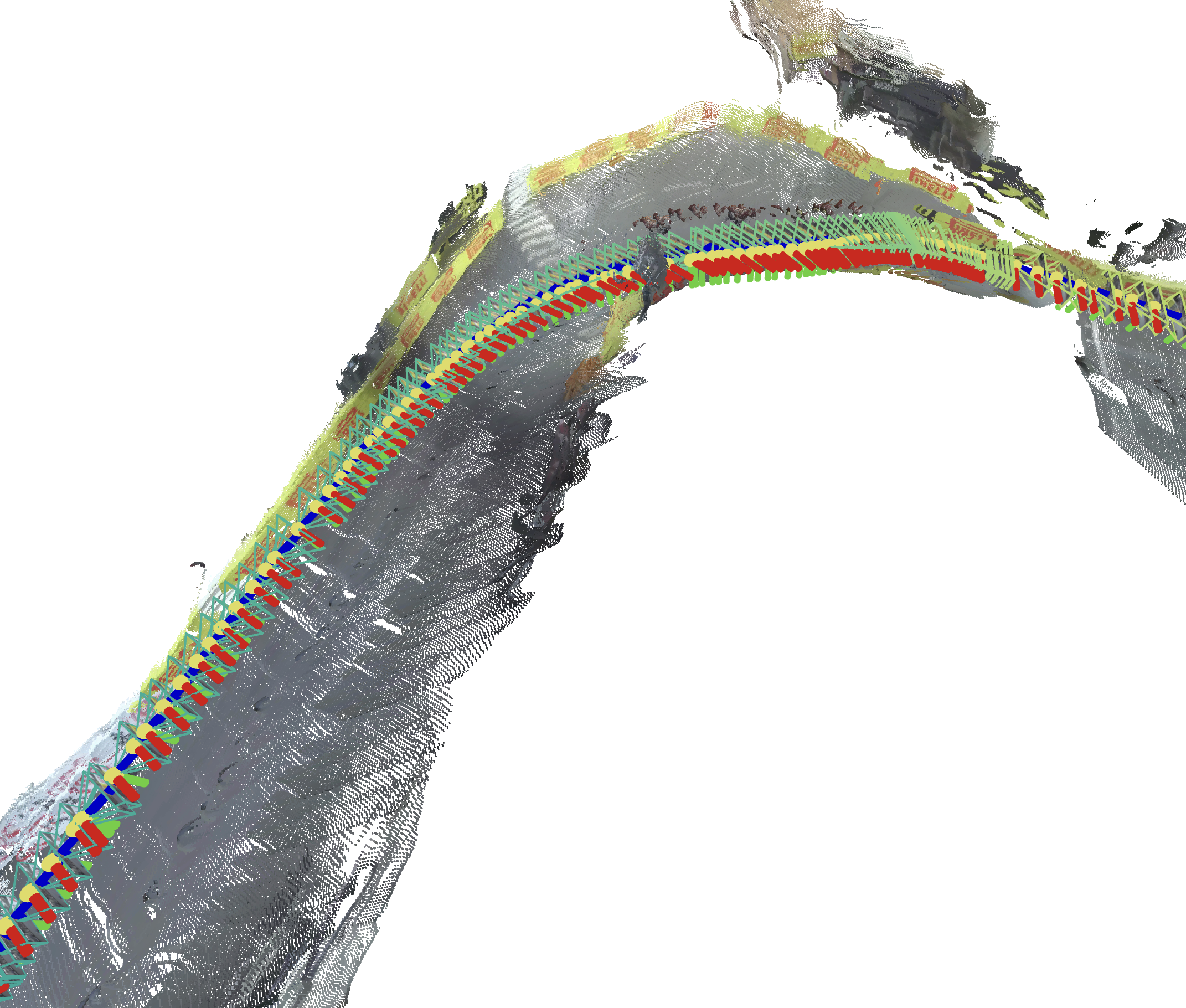} & 
        \includegraphics[width=0.45\linewidth]{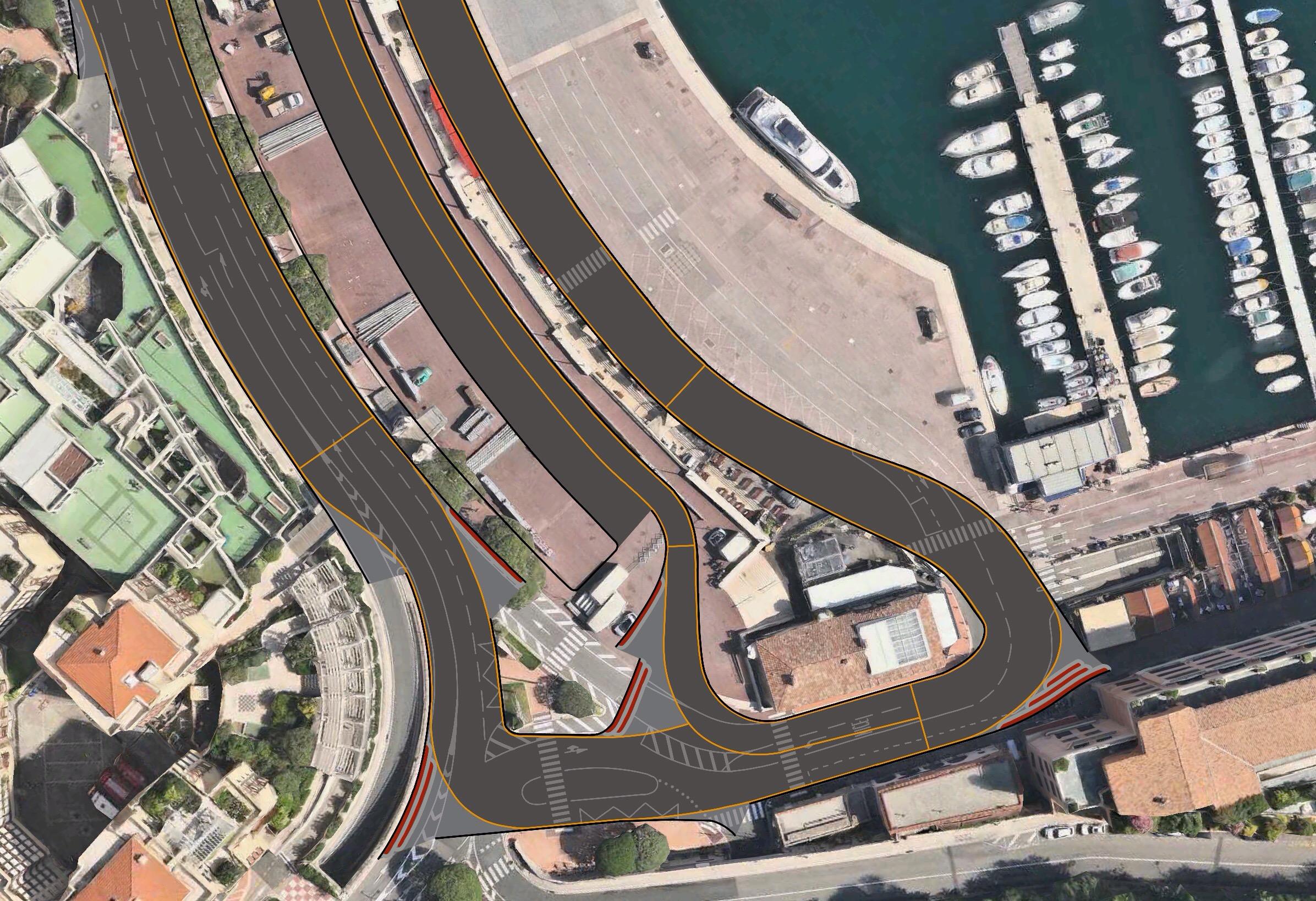} \\
        Turn 18 Reconstruction & Turn 18 Ground Truth \\
        \includegraphics[width=0.45\linewidth]{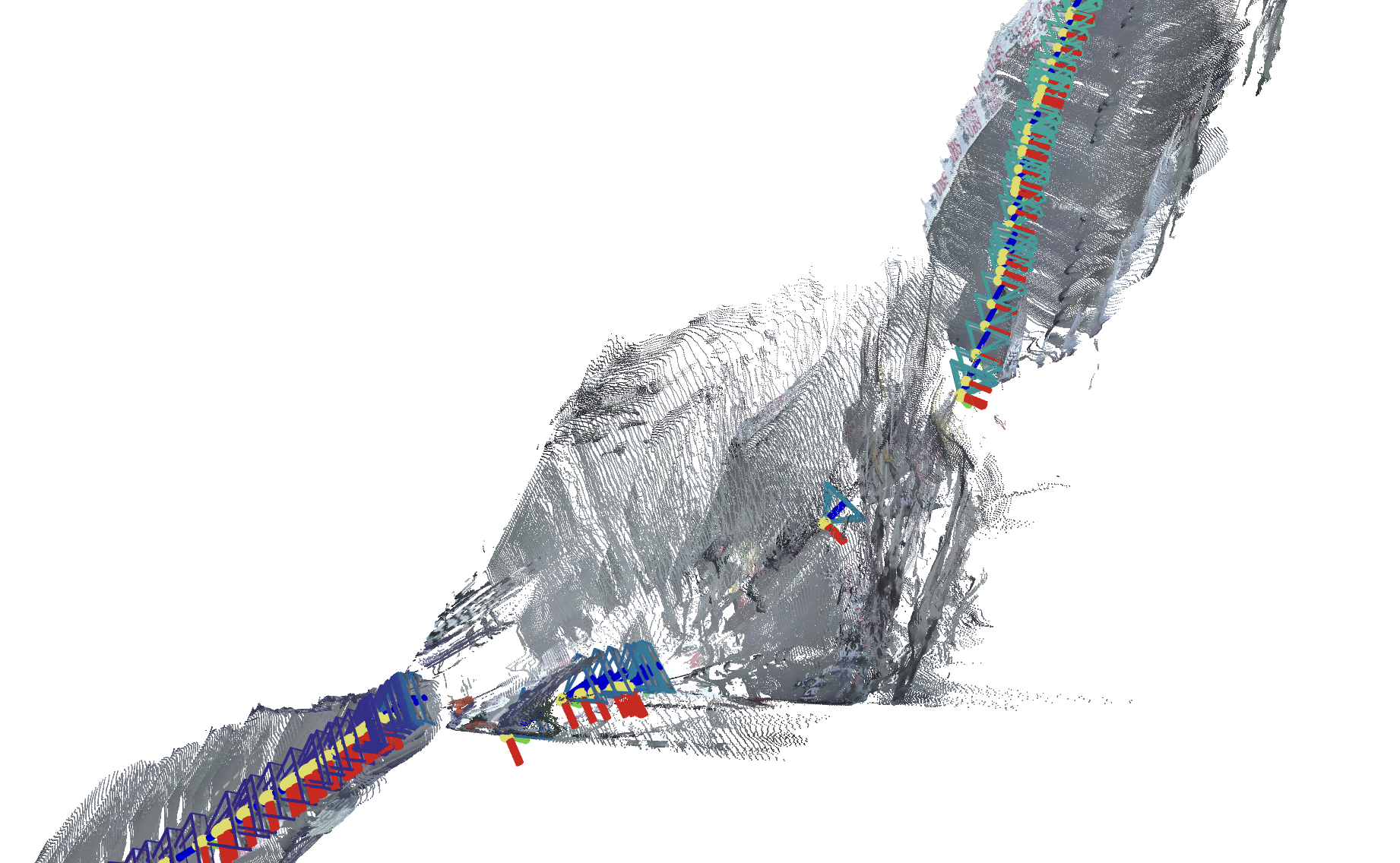} & 
        \includegraphics[width=0.45\linewidth]{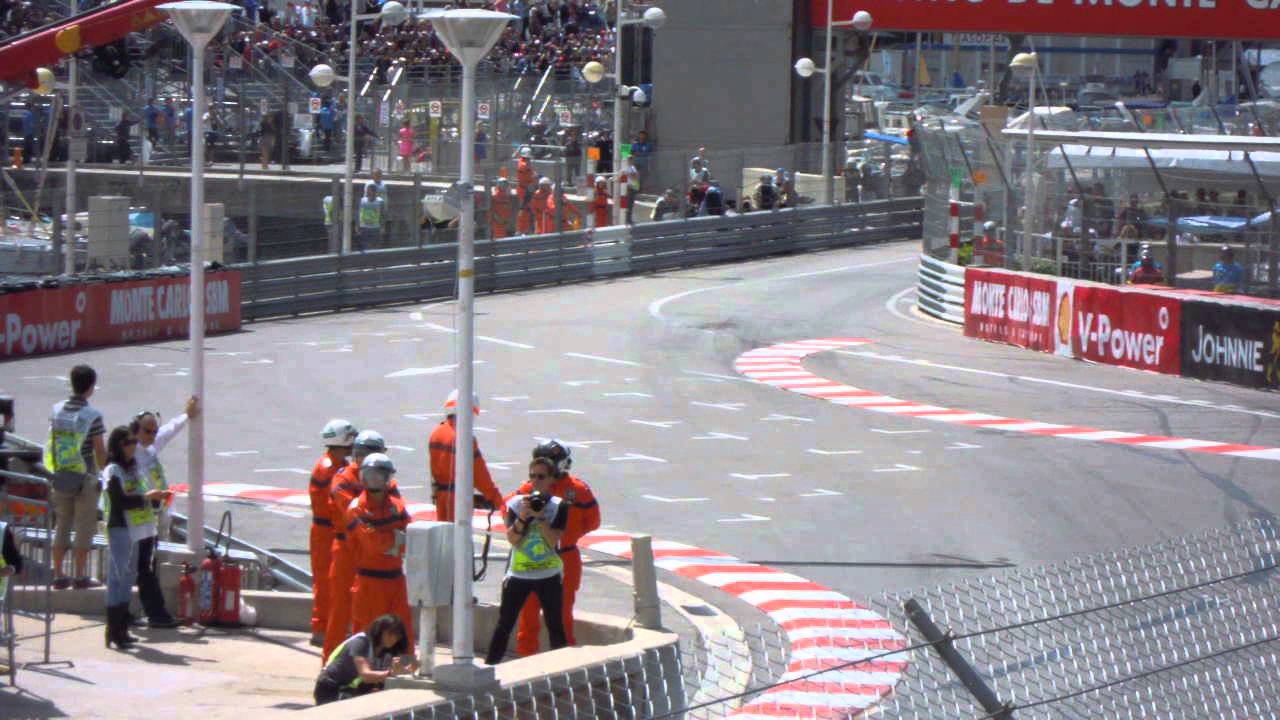} \\
        Failure Case: Turn 15+16 Reconstruction & Turn 15+16 Ground Truth \\
    \end{tabular}
\end{table}

\end{document}